\PassOptionsToPackage{table}{xcolor}
\PassOptionsToPackage{hyphens}{url}
\documentclass[10pt,a4paper]{article}
\usepackage{ant-preprint}
\usepackage[authoryear,round]{natbib}
\usepackage{xspace,array,placeins,float,tabularx,xurl,fvextra,stfloats}
\newcommand{\method}{\mbox{\textsc{CrossPair}}\xspace}
\newcommand{\dataset}{\textsc{Transform-360K}\xspace}
\newcommand{\rs}{x_r}
\newcommand{\rt}{y_r}
\newcommand{\qs}{x_q}
\newcommand{\qt}{y_q}
\newcommand{\Tcal}{\mathcal{T}}
\renewcommand{\PaperTitle}{CrossPair: Unlocking Generalist Visual Transformation Learning in Image Editors}
\renewcommand{\PaperShortTitle}{CrossPair}
\renewcommand{\PaperAuthors}{Yibo Zhao\textsuperscript{1,2},\quad Lifan Jiang\textsuperscript{1,2},\quad Senlin cheng\textsuperscript{2},\quad Tianrun Wu\textsuperscript{3} \\[1pt]
  Bolei Xu\textsuperscript{2},\quad Xiaobo Li\textsuperscript{2},\quad Can Ye\textsuperscript{2} \\[1pt]
  Xiaofei He\textsuperscript{1},\quad Dehai Zhao\textsuperscript{3},\quad Boxi Wu\textsuperscript{3}}
\renewcommand{\PaperAffiliations}{\textsuperscript{1}State Key Lab of CAD\&CG, Zhejiang University\qquad\textsuperscript{2}Ant Group\\[1pt]
  \textsuperscript{3}School of Software Technology, Zhejiang University}
\renewcommand{\PaperHeaderAffiliations}{Zhejiang University\enspace /\enspace Ant Group}
\renewcommand{\PaperContact}{}
\renewcommand{\PaperAuthorNote}{}
\renewcommand{\PaperDate}{October 2026}
\renewcommand{\PaperKeywords}{Visual in-context learning; Image editing; Visual analogy; Exemplar-guided transformation}
\renewcommand{\PaperLinks}{}
\hypersetup{pdfauthor={Yibo Zhao, Lifan Jiang, Senlin cheng, Tianrun Wu, Bolei Xu, Xiaobo Li, Can Ye, Xiaofei He, Dehai Zhao, Boxi Wu},pdfkeywords={CrossPair, visual in-context learning, image editing, visual analogy}}

\renewcommand{\PaperAbstract}{%
Language models switch tasks from demonstrations, but vision still often relies on separate systems for restoration, editing, and stylization.
A before--after example offers a shared instruction only if the transferable change can be separated from the scene that demonstrates it.
We introduce \method, which adapts a pretrained multimodal image editor to transfer diverse transformations from one exemplar pair.
Exemplar-before, exemplar-after, and query images take fixed roles in its native VLM and VAE pathways: the VLM extracts operation semantics, while the VAE retains fine appearance and structure.
A training-only, change-aware DINOv3 objective aligns feature-change directions across content-distinct scenes, guiding the output to follow the demonstrated relation without an inference-time encoder.
We construct \dataset by grouping filtered pairs from conditional representations, restoration, editing, virtual try-on, and stylization under shared transformations.
On Relation252K-derived seen and held-out transformations, \method improves target agreement and edit-direction similarity over released systems.
Ablations show that the semantic and visual exemplar routes are complementary and that DINO relation alignment further strengthens transfer; \dataset training also improves aggregate held-out-operation scores.
}

\begin{document}
\MakeJointTitle

\begin{figure*}[tp]
\centering
\includegraphics[width=\textwidth]{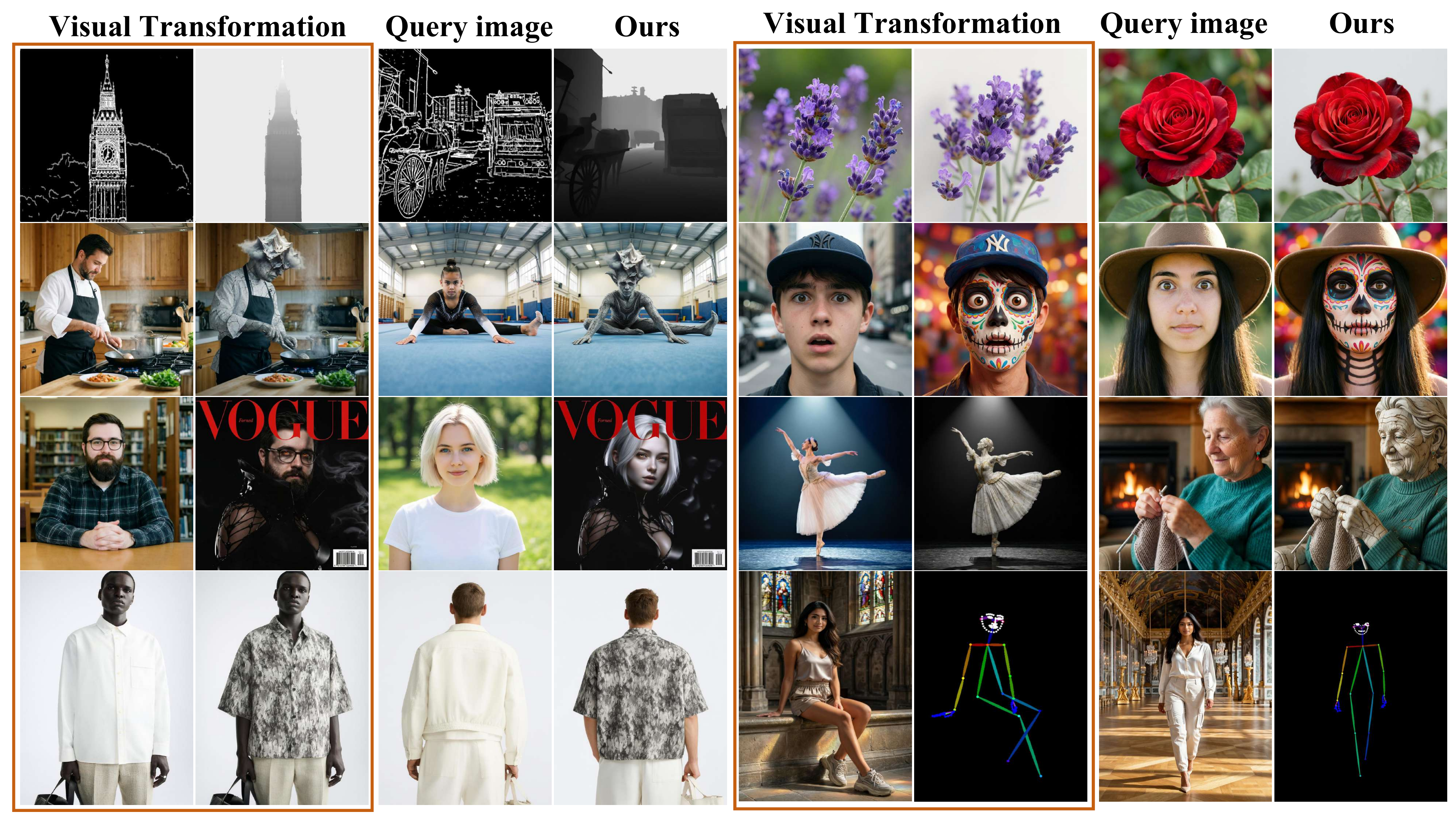}
\caption{Single-pair visual transformation with \method. A before--after exemplar specifies an operation for a content-distinct query.}
\label{fig:teaser}
\end{figure*}

\section{Introduction}
\label{sec:intro}

Many visual transformations are easier to \emph{show} than to name: phrases such as ``make it ceramic'' or ``change the viewpoint'' underspecify texture, geometry, and what should remain untouched.
A before--after pair reveals both the change and the invariant content, offering a common task interface across conditional generation, restoration, editing, try-on, and stylization.
The challenge is to infer that rule from one demonstration and apply it to a different scene.

The demonstrated change is entangled with its reference scene, while a content-distinct query must remain recognizable.
Visual prompting and exemplar-based editing establish pair-conditioned transfer~\citep{visualprompt,imagebrush,edittransfer,relationadapter,lorweb,edittransferpp}; \method coordinates semantic interpretation, fine-grained visual evidence, and explicit relation agreement inside a pretrained multimodal editor.
Given one reference pair and a content-disjoint query, it solves the visual analogy \(\rs : \rt \;::\; \qs : \qt\), where only \((\rs,\rt,\qs)\) are visible at inference.
The pair supplies visual evidence, complemented by record-specific text that can name the edit.

\method realizes this interface through role-ordered dual-space exemplar conditioning.
The reference-before, reference-after, and query images enter a semantic VLM pathway and a visual VAE pathway in the same order.
The VLM interprets the triplet to infer \emph{what operation should transfer}, whereas the VAE preserves the colors, textures, shapes, and local structures needed to reproduce it.
Their fusion combines semantic abstraction with fine-grained visual evidence, while relation supervision encourages the predicted change to follow the demonstrated one.

Conditioning alone, however, does not explicitly require the predicted change to agree with the demonstrated one.
We therefore introduce a training-only, change-aware DINOv3 objective~\citep{dinov3}.
It localizes informative changes independently inside the reference and query scenes, aggregates their patchwise feature displacements into relation directions, and aligns those directions across content.
This formulation avoids assuming correspondence between unrelated scenes and regularizes the demonstrated relation without directly matching prediction and target features.
The complete DINO branch is removed after training.

Finally, we introduce the \dataset corpus, built through six source-specific construction pipelines, hybrid semantic and automatic filtering, and fine-grained relation grouping.
It contains approximately 90K four-image training records (360K record-level image occurrences) spanning over 8K fine-grained relations.
Instantiated by tuning Qwen-Image-Edit-2511, \method provides one model for these heterogeneous operations and improves aggregate target agreement on the reported seen and unseen benchmarks.
Our contributions are:
\begin{itemize}
    \item We adapt a pretrained editor for visual analogy: the exemplar pair and query take fixed roles in native VLM and VAE pathways, joining semantics with fine visual detail.
    \item We introduce a training-only, change-aware DINOv3 objective that aligns reference and predicted-query feature changes across scenes without an extra inference-time encoder.
    \item We build \dataset: 91,010 four-image records spanning more than 8,000 relations, constructed with source-specific filtering and content-distinct pairing by relation.
    \item On seen and held-out tasks, \method improves target agreement and edit-direction transfer; matched ablations isolate VLM/VAE and DINO contributions.
\end{itemize}
\begin{figure*}[tp]
\centering
\includegraphics[width=0.99\textwidth]{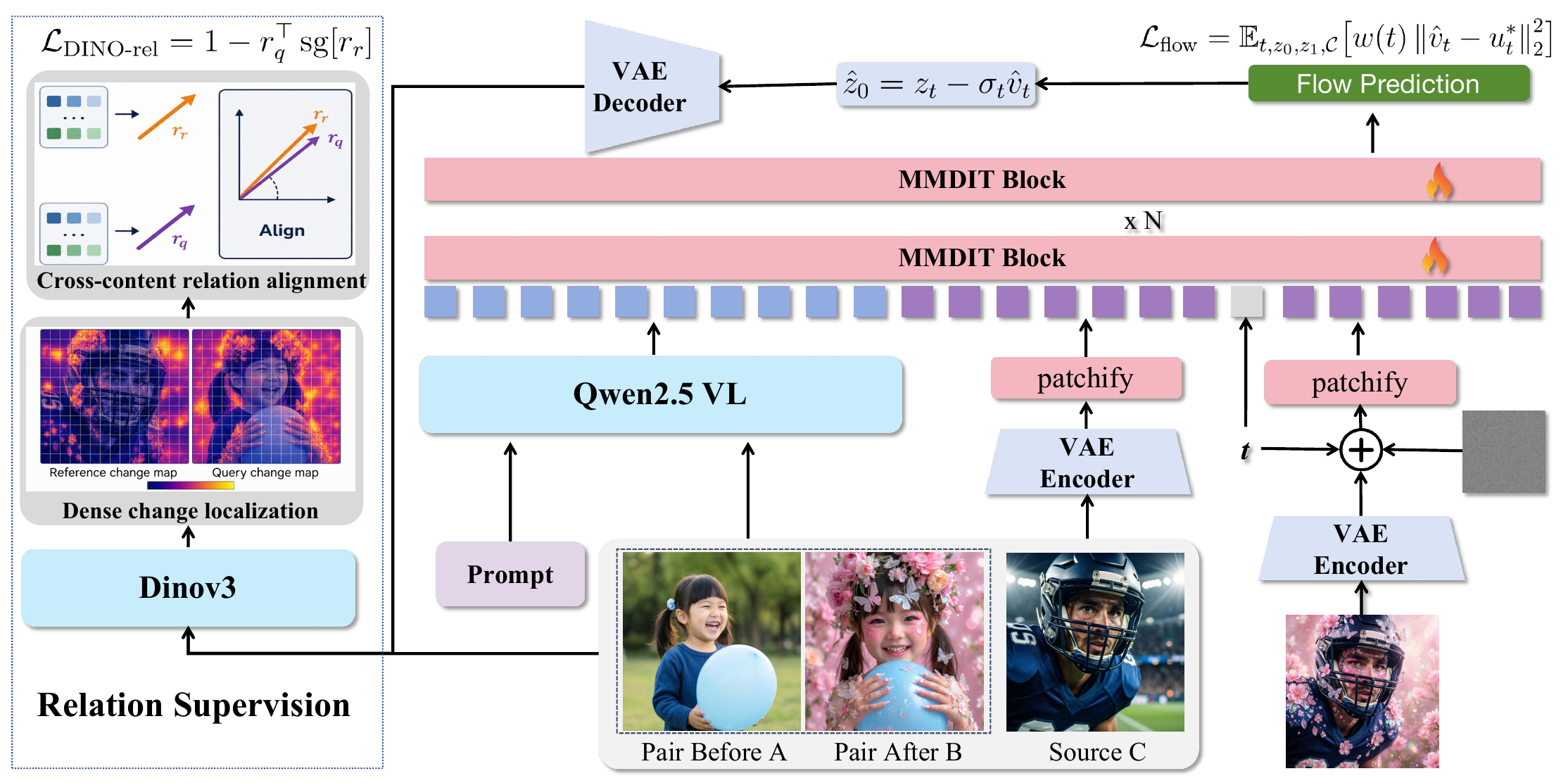}

\caption{Overview of \method.
The reference pair and query jointly enter the native VLM and VAE conditioning routes to guide flow prediction.
During training only, the query target supplies detached change-localization weights and a frozen DINOv3 encoder compares reference and predicted-query relation directions; the VAE/VLM encoders are frozen and only generator LoRA weights are trained.}
\label{fig:method}

\end{figure*}

\section{Related Work}
\label{sec:related}

\subsection{Diffusion-Based Image Editing}
Text-guided editing~\citep{prompttoprompt,instructpix2pix,instructdiffusion,smartedit} and spatial controls~\citep{diffedit,controlnet} expose language or auxiliary maps as task specifications.
Reference-conditioned methods~\citep{paintbyexample,anydoor,ipadapter} transfer appearance or object content.
A before--after exemplar instead demonstrates a change.
Qwen-Image-Edit~\citep{qwenimage} already offers semantic VLM and reconstructive VAE conditioning; \method assigns exemplar and query roles in these existing routes and uses a frozen DINOv3 branch only during training.

\subsection{Visual In-Context Editing}
Image analogies~\citep{imageanalogies}, Visual Prompting~\citep{visualprompt}, Painter~\citep{painter}, and ImageBrush~\citep{imagebrush} established visual demonstration as a task interface.
PairEdit~\citep{pairedit}, Edit Transfer~\citep{edittransfer}, RelationAdapter~\citep{relationadapter}, LoRWeB~\citep{lorweb}, VIRAL~\citep{viral}, and EditTransfer++~\citep{edittransferpp} all address exemplar-based editing; Textualize Visual Prompt~\citep{textualizevisualprompt} translates pair evidence into text through a diffusion bridge, while EditCLIP~\citep{editclip} learns pair-level edit representations.
Thus one-pair transfer itself is established.
Compared with per-exemplar adaptation and dedicated relation encoders, our contribution is a fixed triplet interface in the pretrained editor's two native pathways plus change-weighted relation regularization with no inference-time auxiliary encoder.
Table~\ref{tab:ablation} studies the proposed components within a matched backbone.

\section{Method}
\label{sec:method}

\subsection{Rectified Flow}
\label{sec:rectified-flow}

Our model builds upon Qwen-Image-Edit~\citep{qwenimage}, which formulates image generation as a rectified-flow process in latent space~\citep{rectifiedflow,flowmatching}.
Let \(z_0\) denote a clean target latent and \(z_1\sim\mathcal{N}(0,I)\) a noise latent.
For interpolation coefficient \(\sigma_t\in[0,1]\), rectified flow follows the straight path
\begin{equation}
    z_t=(1-\sigma_t)z_0+\sigma_t z_1,
    \qquad u_t^*=z_1-z_0.
    \label{eq:rf-path}
\end{equation}
A velocity network \(v_\theta\) predicts this constant target velocity from the noisy latent, timestep, and condition \(\mathcal{C}\).
We denote its output by \(\hat v_t=v_\theta(z_t,t;\mathcal{C})\) and train it with
\begin{equation}
    \mathcal{L}_{\mathrm{flow}}
    =
    \mathbb{E}_{t,z_0,z_1,\mathcal{C}}\!\left[
    w(t)\left\|
    \hat v_t-u_t^*
    \right\|_2^2\right].
    \label{eq:flow-loss}
\end{equation}
Here \(w(t)\) is the fixed training weight supplied by the Qwen flow scheduler, not a learned relation weight; timesteps are sampled over the configured training interval.
The predicted velocity also yields a differentiable one-step estimate of the clean latent:
\begin{equation}
    \hat z_0=z_t-\sigma_t\hat v_t.
    \label{eq:clean-pred}
\end{equation}

\subsection{One-Pair Inference, Cross-Pair Supervision}
\label{sec:formulation}

Let \(\Tcal\) denote a transformation.
A training item contains one reference pair \((\rs,\rt)\), where \(\rt=\Tcal(\rs)\), and one query pair \((\qs,\qt)\), where \(\qt=\Tcal(\qs)\).
At inference, the model receives \((\rs,\rt,\qs)\) and optional text \(p\), and predicts
\begin{equation}
    \hat\qt = G_\theta(\rs,\rt,\qs;p).
    \label{eq:problem}
\end{equation}
There is one demonstration pair per forward pass and at inference.
However, the aligned query target makes the \emph{training unit} a pair of pairs.
We call the setting one-pair prompting, not single-pair supervision; the latter describes methods that self-supervise from \((\rs,\rt)\) alone~\citep{deltaadapter}.
The evaluated Relation252K records combine a role instruction with a concrete edit description when available; one such description requests teeth alignment and whitening.
Appendix~\ref{app:prompt-evidence} records the serialized prompt.

The central modeling decision is how to expose the exemplar.
We represent the same ordered triplet in two complementary spaces: a fine-grained VAE condition \(C_{\mathrm{vae}}\) and a semantic VLM condition \(C_{\mathrm{vlm}}\).
They are not competing transformation encoders; together they preserve what the change looks like and explain what operation it expresses.

\subsection{Role-Ordered Exemplar Conditioning}
\label{sec:conditioning}

Qwen-Image-Edit normally represents a source image through both a reconstructive VAE path and a semantic VLM path.
We extend both native paths to the ordered triplet \((\rs,\rt,\qs)\).
Let \(E_{\mathrm{vae}}\) denote the frozen VAE encoder, \(P\) its latent patchification, and \(E_{\mathrm{vlm}}\) the frozen Qwen2.5-VL encoder.
Our two conditions are
\begin{equation}
\begin{aligned}
    C_{\mathrm{vae}}
    &=\operatorname{cat}\!\left\{
      P(E_{\mathrm{vae}}(I)): I\in[\rs,\rt,\qs]\right\},\\
    C_{\mathrm{vlm}}
    &=E_{\mathrm{vlm}}([\rs,\rt,\qs],p).
\end{aligned}
    \label{eq:conditions}
\end{equation}
The shared order assigns fixed roles---\emph{reference before}, \emph{reference after}, and \emph{query}---in both branches.
Joint VLM states abstract the operation in the context of the new source, whereas VAE tokens retain precise color, texture, and local structure.
VLM-only conditioning may compress appearance-critical evidence; VAE-only conditioning preserves that evidence but leaves the transformation entangled with reference content.
Their fusion provides both abstraction and fidelity, while the relation objective in Sec.~\ref{sec:relation} explicitly regularizes which change should transfer.
No new trainable or inference-time encoder is introduced; a frozen DINOv3 encoder is used only during training, and only DiT LoRA weights are adapted.

For our editor, \(z_0=E_{\mathrm{vae}}(\qt)\) and \(\mathcal{C}=(C_{\mathrm{vae}},C_{\mathrm{vlm}})\) in Eq.~\eqref{eq:flow-loss}.
Only the query target is noised and scored.

\subsection{Change-Aware DINO Relation Alignment}
\label{sec:relation}

\paragraph{Why endpoint supervision is insufficient.}
Flow matching supervises the complete target as an endpoint, but visual analogy additionally requires equality between two \emph{changes}:
\begin{equation}
    \mathcal{R}(\rs,\rt)
    \simeq
    \mathcal{R}(\qs,\hat\qt),
    \qquad \rs\not\approx\qs,
    \label{eq:relation-goal}
\end{equation}
where \(\mathcal{R}\) should retain the demonstrated operation while discarding scene identity.
This cannot be implemented by pixel differences or direct patch matching because the reference and query depict unrelated content, and unchanged regions can dominate a global feature average.
We therefore construct \(\mathcal{R}\) in two stages: first identify informative changes independently inside each scene, then compare the aggregated change directions across scenes.
To obtain a differentiable image prediction, we decode Eq.~\eqref{eq:clean-pred} as
\(\hat\qt=D_{\mathrm{vae}}(\hat z_0)\) and extract all relation features with a frozen DINOv3 encoder~\citep{dinov3}.
The VAE decoder is frozen but remains differentiable with respect to \(\hat z_0\).

\paragraph{Dense change localization.}
Our frozen DINOv3 ViT-7B/16 encoder yields normalized spatial features
\(\bar f_i(I)\) on a \(14\!\times\!14\) patch grid.
A uniform average would be dominated by unchanged background.
Let \((I_r^-,I_r^+)=(\rs,\rt)\) and
\((I_q^-,I_q^+)=(\qs,\qt)\).
We estimate change in either scene \(d\in\{r,q\}\) with one shared definition:
\begin{equation}
\begin{aligned}
 c_i^d&=1-\bar f_i(I_d^-)^\top\bar f_i(I_d^+),\\
 w_i^d&=\frac{(c_i^d)^\alpha}{\sum_j(c_j^d)^\alpha+\epsilon},
 \quad d\in\{r,q\}.
\end{aligned}
\label{eq:change-weights}
\end{equation}
where \(\alpha=1\) in all experiments.
The fixed \(\epsilon>0\) stabilizes empty or near-empty change maps; all feature maps are computed on the same grid within each before--after pair.
Each \(c_i\) is a same-location feature displacement computed only between the before and after views of one scene; a larger value indicates that the patch is more informative about the transformation.
The normalized weights act as soft change maps.
Reference and query maps are estimated independently and may attend to different spatial locations, which helps when the same operation acts on different objects or layouts.
The query target determines the detached weights \(w^q\), so target information enters the training loss through localization.
No target-feature gradient or direct target-to-prediction feature matching is used in the relation term; the query target is unavailable at inference.

\paragraph{Cross-content relation alignment.}
For relation extraction, set
\(\widetilde I_r^+=\rt\) and \(\widetilde I_q^+=\hat\qt\).
We aggregate either scene's patchwise displacement with the same expression:
\begin{equation}
\begin{gathered}
r_d=\operatorname{norm}\!\left(
  \sum_i w_i^d
  [\bar f_i(\widetilde I_d^+)-\bar f_i(I_d^-)]\right),\\
d\in\{r,q\}.
\end{gathered}
\label{eq:dino-relations}
\end{equation}
Here \(\operatorname{norm}(v)=v/(\|v\|_2+\epsilon)\), with the same fixed stabilizer as above.
The weighted sum converts a spatially distributed edit into one scene-level transformation vector.
Normalization removes overall response magnitude, so the objective compares \emph{how features move} rather than how much activation a particular scene produces.
Patch indices are paired only within the same scene: \(\rs\) with \(\rt\), and \(\qs\) with \(\qt\) or \(\hat\qt\).
We never assume that patch \(i\) in the exemplar corresponds to patch \(i\) in the unrelated query.
Only after within-scene aggregation do we align the global directions:
\begin{equation}
    \mathcal{L}_{\mathrm{DINO\text{-}rel}}
    =1-r_q^\top\operatorname{sg}[r_r].
    \label{eq:dino-relation-loss}
\end{equation}
DINO branches are frozen; only \(\bar f(\hat\qt)\) retains its input gradient.
Eq.~\eqref{eq:dino-relation-loss} never matches prediction features directly to target features; it transfers the exemplar's change direction.

\paragraph{Reliable low-noise supervision.}
At high noise, a one-step \(\hat\qt\) is too uncertain for structural supervision.
We use
\begin{equation}
    g(\sigma_t)=\max\!\left(0,
    \frac{\sigma_{\max}-\sigma_t}{\sigma_{\max}}\right),
    \label{eq:sigma-weight}
\end{equation}
and optimize
\begin{equation}
    \mathcal{L}
    =\mathcal{L}_{\mathrm{flow}}
    +\lambda_{\mathrm{dino}}\,
    \mathbb{E}\!\left[g(\sigma_t)
    \mathcal{L}_{\mathrm{DINO\text{-}rel}}\right].
    \label{eq:total-loss}
\end{equation}
The gate threshold and loss weight are given in the supplementary material.
No direct perceptual, same-source delta, or auxiliary VLM relation loss is active.

\paragraph{Training setup.}
The VLM, VAE, base generator, and DINOv3 encoder remain frozen; optimization updates rank-128 LoRA adapters in the image generator.
Each step samples one four-image record and one flow timestep, using the native scheduler weight for the primary MSE objective.
We train with AdamW at a constant \(10^{-4}\) learning rate, bfloat16 precision, and an aspect-preserving image area cap of 262,144 pixels; Appendix~\ref{app:training} gives the full optimizer and preprocessing configuration.

\subsection{\dataset}
\label{sec:dataset-construction}
\label{sec:large-dataset}

\begin{table*}[tp]
\caption{Source composition of the frozen Transform-360K training manifest.}
\label{tab:corpus-composition}
\centering
\small
\setlength{\tabcolsep}{3pt}
\begin{tabularx}{\textwidth}{@{}Xrr@{\hspace{1.5em}}Xrr@{}}
\toprule
Manifest segment & Records & Share & Manifest segment & Records & Share \\
\midrule
Low-level vision & 20,049 & 22.03\% & Style transfer sources & 15,039 & 16.53\% \\
Relation252K & 17,600 & 19.34\% & Virtual try-on & 13,527 & 14.86\% \\
Conditional representations & 11,645 & 12.80\% & Reference and segmentation & 10,443 & 11.47\% \\
Spatial transformations & 2,144 & 2.36\% & Other paired sources & 563 & 0.62\% \\
\midrule
\multicolumn{6}{@{}l@{}}{\textbf{Total:} 91,010 records (100\%)} \\
\bottomrule
\end{tabularx}

\end{table*}

\begin{figure*}[tp]
\centering
\includegraphics[width=\textwidth]{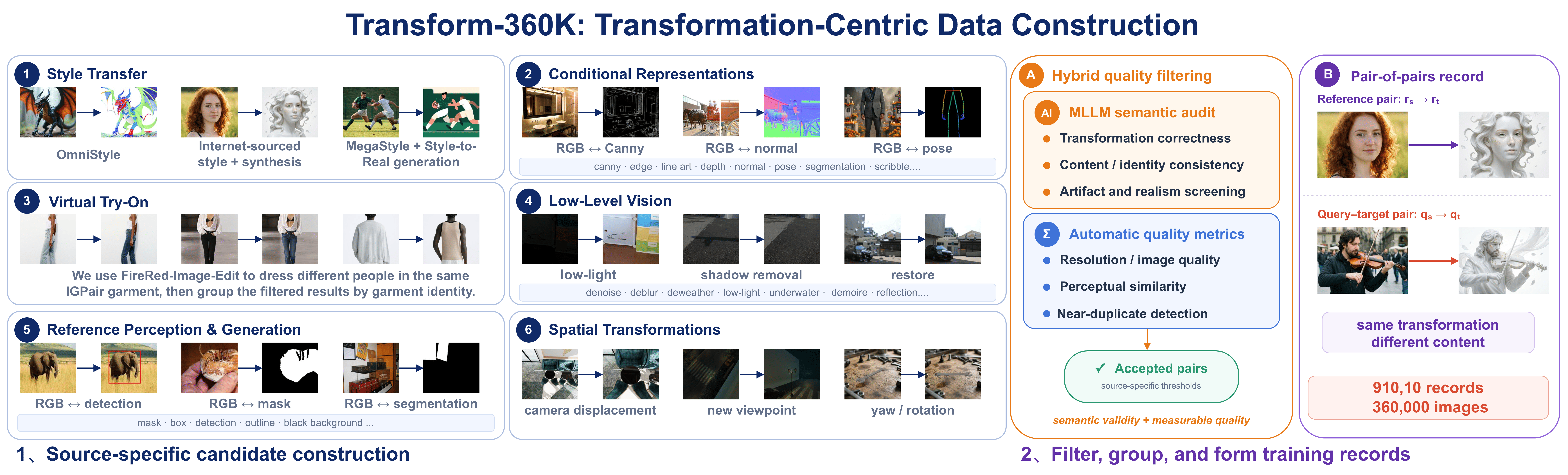}
\caption{Transform-360K construction: source-specific pairing, quality filtering, and cross-pair formation within operation keys.}
\label{fig:dataset}
\end{figure*}

\paragraph{Transformation canonicalization.}
We do not pool heterogeneous before--after images under dataset names or broad task labels.
Each candidate is normalized as \((x,y,g)\), where \(g\) records the finest verifiable operation and its direction.
For style and try-on, \(g\) is the style or garment identity~\citep{omnistyle,megastyle,igpair}; for conditional representations it is an ordered modality mapping (e.g., RGB\(\rightarrow\)depth is distinct from depth\(\rightarrow\)RGB); for restoration it is the degradation task; and for spatial edits it is a discretized viewpoint, distance, or rotation delta~\citep{spatialedit}.
The six construction streams instantiate these keys from style pairs, RGB/edge/depth/normal/pose/line-art/mask/box/segmentation mappings, garment transfers, low-level restoration, reference-driven perception\slash generation, and controlled geometry.

\paragraph{Pair validation.}
Every pair must pass common checks for readability, resolution, duplicate status, and a non-trivial but bounded visual change.
An MLLM then verifies that the claimed operation is visible, non-target content remains consistent, and artifacts do not explain the difference.
Stream-specific predicates make this test concrete: style pairs reject content drift or residual target style in the source; condition maps reject empty or invalid structure; try-on pairs require the same garment on different people; restoration pairs retain the same scene and declared degradation; and spatial pairs must realize the keyed geometric delta.
Conventional image and perceptual statistics screen obvious failures before the semantic audit.

\paragraph{Construction audit.}
Candidates pass full decoding and geometry preflight before relation assembly.
In one captioned-source pass, 67,271 of 72,273 records survive these checks; a separate MegaStyle semantic audit retains 7,552 of 12,450 candidate pairs when both images pass its consistency test.
The retained pairs then enter operation-specific validation and cross-content pairing, with the detailed source inventory in Appendix~\ref{app:dataset}.

\paragraph{Cross-pair record formation.}
Within each key \(g\), we require distinct content/scene identities, reject self- or shared-scene pairings, and sample one pair as \((\rs,\rt)\) and another as \((\qs,\qt)\).
We then balance relation frequency and source families rather than allowing large collections to dominate.
The result contains approximately 90K four-image records (360K record-level image occurrences) across over 8K fine-grained relations, compared with the 200+ transformation groups in Relation252K.
The supplementary material provides source inventories, the serialized record schema, filtering criteria, split construction, and leakage-audit details.

\paragraph{Corpus composition.}
The frozen manifest has 91,010 training records, each with a before--after exemplar and an aligned query pair.
Table~\ref{tab:corpus-composition} summarizes its source mix; the style sources are grouped for display, while the full manifest breakdown appears in Appendix~\ref{app:dataset}.
The imported Relation252K segment provides common-operation coverage, and the remaining streams contribute conditional representations, try-on, restoration, reference tasks, style transfer, and spatial edits.
The common assembler balances groups after the source-specific validation described above, so each training record expresses one operation across distinct content.

\section{Experiments}
\label{sec:experiments}

\subsection{Evaluation Setup}
\label{sec:relation252k}

\paragraph{Training-data protocol and fairness.}
We separate the core method comparison from the dataset-scaling study.
RelationAdapter~\citep{relationadapter} is trained on Relation252K, whereas the released LoRWeB model~\citep{lorweb} uses Relation252K together with additional training data.
Our primary \method model uses a 265K training mixture derived from Relation252K, and the \dataset checkpoint is evaluated separately.
A second checkpoint trained on \dataset is added only to the OOD panel.

\paragraph{Baselines and inference controls.}
We compare with the publicly available LoRWeB~\citep{lorweb} and RelationAdapter~\citep{relationadapter} systems, together with Qwen-Image-Edit-2511~\citep{qwenimage}.
All methods receive the same held-out record and one exemplar pair; extracted outputs are resized to the same metric resolution.
Released systems retain their native inference procedures, so this is a complete-system comparison rather than a mechanism-matched backbone study.

\paragraph{Evaluation protocol.}
For in-domain, we test 219 seen transformation groups with five held-out queries each.
For out-of-domain, we augment Relation252K's released unseen-task split with 150 additional cases from diverse editing families; unseen denotes held-out transformation groups, not merely new images within a seen group.
The resulting protocol comprises 1,095 seen-task queries and 450 OOD queries; query records are excluded from their corresponding training groups and cannot serve as their own demonstrations.
We report CLIP-I against the ground-truth target and query source, LPIPS against the target, and CLIP directional similarity.
For GPT-5.5 evaluation, the judge first infers the transformation from the exemplar pair and then ranks the anonymized candidate outputs (four in-domain and five OOD).
The overall ranking prioritizes transformation faithfulness, followed by query-content preservation and visual quality; two diagnostic rankings isolate the first two criteria.
Neither the ground-truth target nor method identities are shown to the judge.
We report the single-judge rankings as diagnostic preferences and the mixed-family CLIP/LPIPS scores as benchmark averages.

\subsection{Results}
\label{sec:main-comparison}

\begin{table*}[t]
\captionsetup{skip=5pt}
\caption{In-domain and OOD metrics with single-run GPT-5.5 Rank-1 rates. R252K and T360K denote the two training mixtures.}
\label{tab:relation252k-main}
\centering
\small
\setlength{\tabcolsep}{3pt}
\begin{tabularx}{\textwidth}{@{}Xrrrrrrr@{}}
\toprule
& \multicolumn{4}{c}{Deterministic metrics}
& \multicolumn{3}{c}{GPT-5.5 Rank-1 rate (\%)} \\
\cmidrule(lr){2-5}\cmidrule(lr){6-8}
Method & CLIP-I$_{\mathrm{GT}}\uparrow$ & CLIP-I$_{\mathrm{src}}\uparrow$
& CLIP-Dir$\uparrow$ & LPIPS$_{\mathrm{GT}}\downarrow$
& Overall & Trans. & Content \\
\midrule
\multicolumn{8}{@{}l}{\textbf{(a) In-domain: 219 seen transformation groups}} \\
\addlinespace[2pt]
LoRWeB~\citep{lorweb} & 0.772 & \textbf{0.690} & 0.309 & 0.479
& 17.3 & 15.1 & \textbf{47.8} \\
RelationAdapter~\citep{relationadapter} & 0.777 & 0.602 & 0.355 & 0.429
& 24.0 & 21.3 & 22.3 \\
Qwen Edit~\citep{qwenimage} & 0.715 & 0.599 & 0.302 & 0.614
& 5.0 & 4.0 & 14.7 \\
\midrule
\rowcolor{AntBlue!8}\method (R252K) & \textbf{0.871} & 0.572 & \textbf{0.438} & \textbf{0.361}
& \textbf{53.7} & \textbf{59.6} & 15.3 \\
\midrule
\multicolumn{8}{@{}l}{\textbf{(b) Out-of-domain: unseen transformations}} \\
\addlinespace[2pt]
LoRWeB & 0.717 & \textbf{0.774} & 0.230 & 0.480
& 10.5 & 11.0 & 29.0 \\
RelationAdapter & 0.722 & 0.736 & 0.253 & 0.476
& 12.4 & 9.8 & \textbf{32.2} \\
Qwen Edit & 0.683 & 0.645 & 0.261 & 0.538
& 6.5 & 6.3 & 13.8 \\
\midrule
\rowcolor{AntBlue!8}\method (R252K) & 0.744 & 0.674 & 0.312 & 0.440
& 27.1 & 28.7 & 13.8 \\
\rowcolor{AntBlue!8}\method (T360K) & \textbf{0.783} & 0.642 & \textbf{0.370} & \textbf{0.402}
& \textbf{43.5} & \textbf{44.2} & 11.2 \\
\bottomrule
\end{tabularx}
\end{table*}

\begin{figure*}[!t]
\centering
\includegraphics[width=0.84\textwidth]{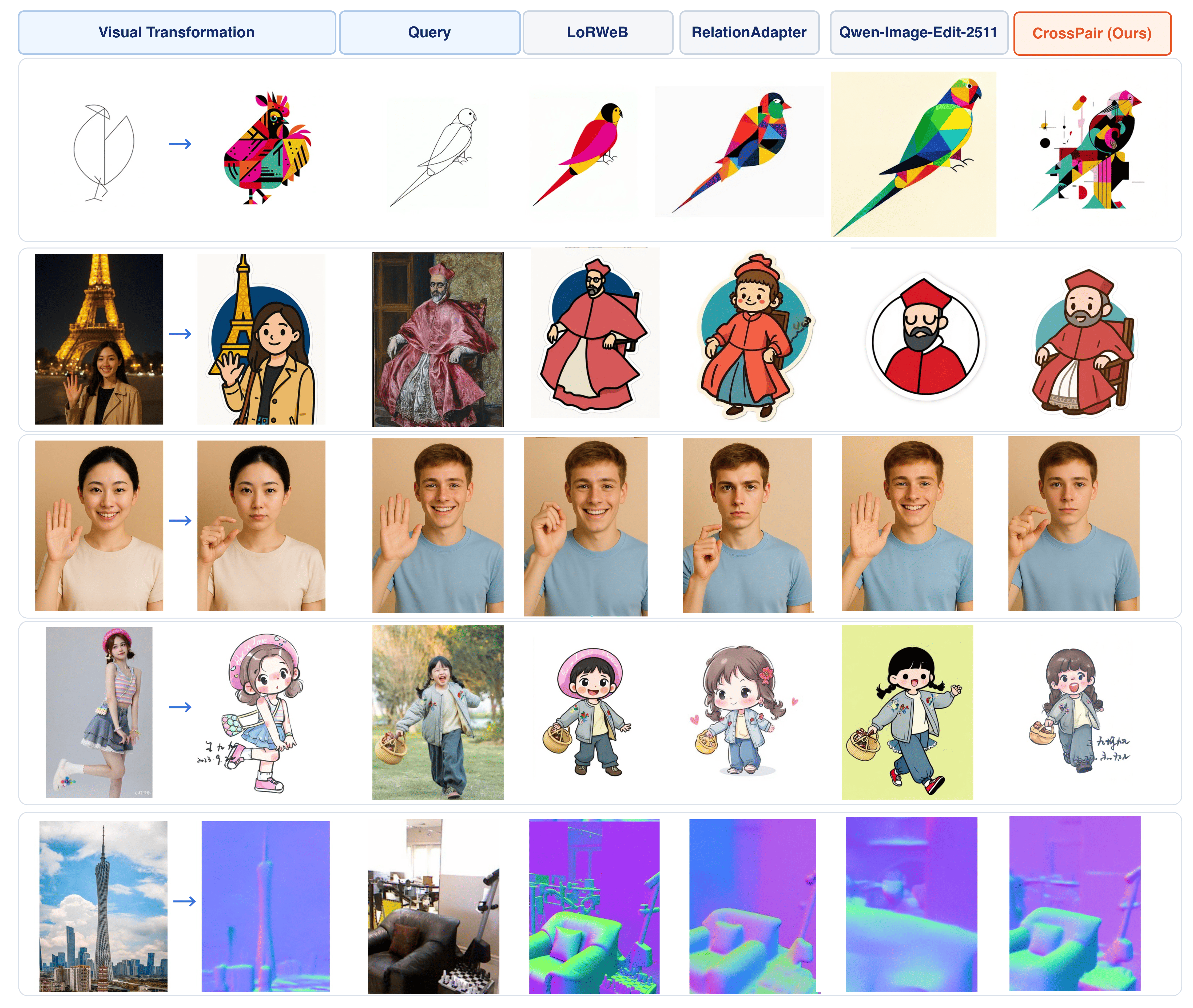}

\caption{We compare \method with LoRWeB, RelationAdapter, and Qwen-Image-Edit-2511 on in-domain Relation252K.}
\label{fig:qual-id}

\end{figure*}

\begin{figure*}[t]
\captionsetup{skip=5pt}
\centering
\includegraphics[width=0.78\textwidth]{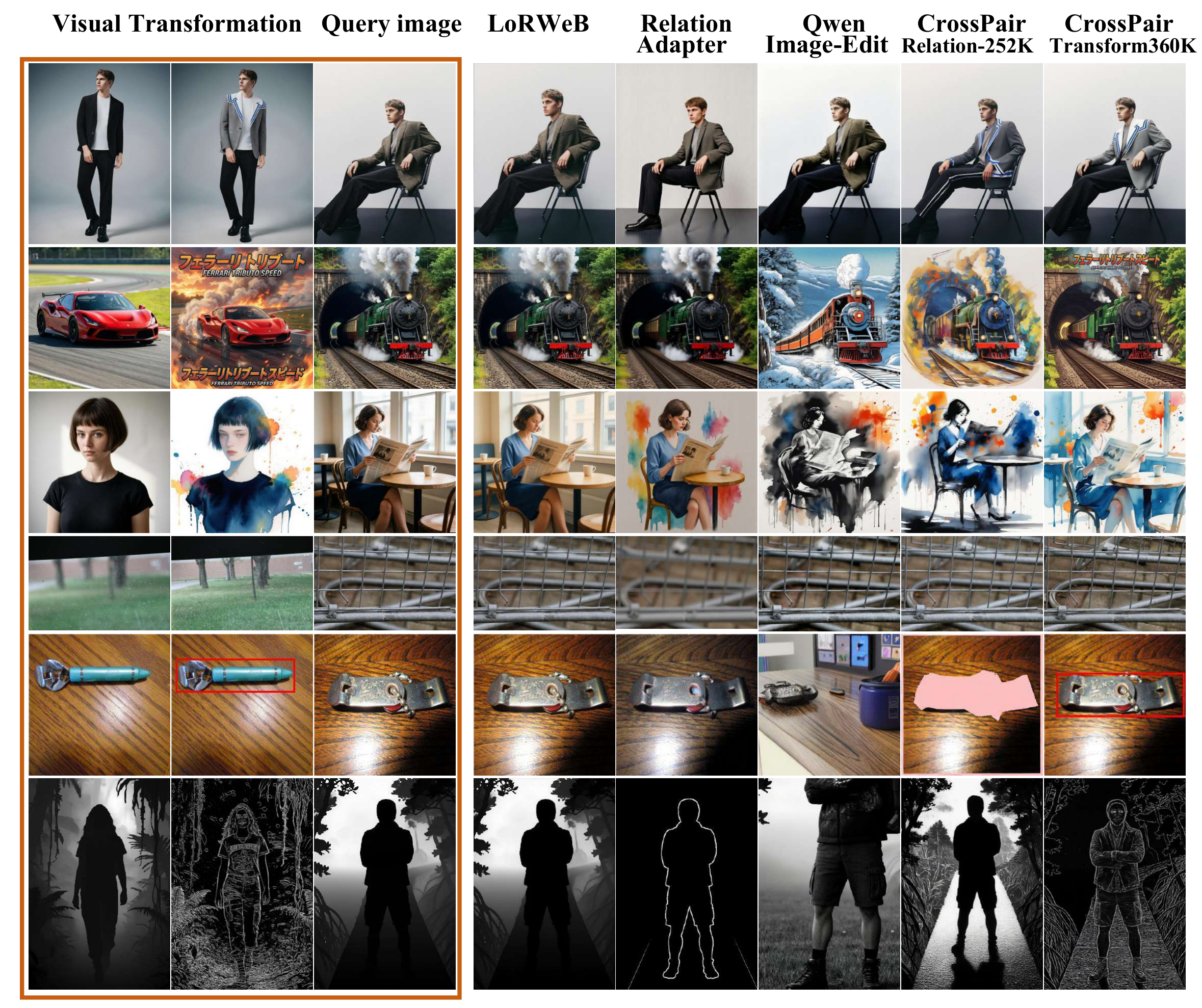}
\caption{OOD comparison of Relation252K- and Transform-360K-trained CrossPair with released baselines.}
\label{fig:qual-ood}
\end{figure*}

\paragraph{Quantitative evaluation.}
The reported measures separate target agreement, operation transfer, and query retention. CLIP-I$_{\mathrm{GT}}$ and LPIPS$_{\mathrm{GT}}$ compare each prediction with its held-out target; CLIP-Dir tests whether the query change follows the demonstration; CLIP-I$_{\mathrm{src}}$ and the content ranking probe retained query content. Reading these views together makes the transformation--preservation balance explicit for the methods in Table~\ref{tab:relation252k-main}.
Without using \dataset, \method has the highest in-domain target agreement and edit-direction similarity in Table~\ref{tab:relation252k-main}.
LoRWeB leads on source similarity and content-preservation preference, whereas \method emphasizes transformation transfer; its content Rank-1 rates are 15.3\% in-domain and 11.2\% for the \dataset OOD model.
The two checkpoints serve different comparisons. The Relation252K-derived model isolates the method on the common-operation benchmark, while the \dataset checkpoint evaluates the same interface with a broader operation inventory. The latter improves OOD aggregate target and directional scores, reflecting the combined change in corpus size and composition. In both splits, target agreement and directional similarity rise together, consistent with transferring the demonstrated change rather than only producing a plausible target image.
GPT-5.5 rankings likewise favor \method on transformation faithfulness.

\paragraph{Qualitative evaluation.}
Figures~\ref{fig:qual-id} and~\ref{fig:qual-ood} compare the methods on in-domain and out-of-domain transformations, respectively.
The primary comparison uses the Relation252K-trained \method so that our new corpus cannot explain its gains; the OOD figure additionally includes the \dataset-trained variant as a separate generalization study.
Across both settings, we examine whether a method transfers the demonstrated change without copying exemplar content or unnecessarily altering the query.

\subsection{Ablation Study}
\label{sec:ablation}

\begin{table}[H]
\captionsetup{skip=5pt}
\caption{OOD component ablation; the final two rows isolate the DINO relation loss.}
\label{tab:ablation}
\centering
\footnotesize
\setlength{\tabcolsep}{2.2pt}
\begin{tabular}{@{}lrrrr@{}}
\toprule
\multicolumn{5}{c}{\textbf{OOD component ablation on Relation252K}} \\
\midrule
Configuration & \shortstack{CLIP-I\\$\mathrm{GT}\uparrow$} & \shortstack{CLIP-I\\$\mathrm{src}\uparrow$} & \shortstack{CLIP-Dir\\$\uparrow$} & \shortstack{LPIPS\\$\mathrm{GT}\downarrow$} \\
\midrule
VAE exemplar & 0.711 & \textbf{0.685} & 0.279 & 0.426 \\
VLM exemplar & 0.768 & 0.618 & 0.343 & 0.406 \\
VAE+VLM exemplar & 0.776 & 0.620 & 0.350 & 0.402 \\
\rowcolor{AntBlue!8}\quad + DINO relation & \textbf{0.778} & 0.622 & \textbf{0.369} & \textbf{0.400} \\
\bottomrule
\end{tabular}

\end{table}

All variants use Relation252K and matched optimization on its unseen-task OOD split. Each retains the query's standard VAE/VLM routes; row labels indicate which route also receives the exemplar. The last two rows isolate DINO relation supervision under joint conditioning.

The VAE-only route has the highest source similarity, reflecting the role of fine visual evidence in retaining the query. The VLM-only route yields stronger target and directional scores than the VAE-only route, consistent with its semantic view of the demonstrated operation. Combining both routes raises target agreement beyond either alone. With those routes held fixed, the DINO relation objective gives the strongest directional score and lowest target LPIPS in Table~\ref{tab:ablation}. This pattern supports the intended division of labor: native editor pathways carry the exemplar at inference, while change-aware supervision shapes the transferred relation during training.

\section{Conclusion}
\label{sec:conclusion}

\method turns a pretrained editor into a one-pair transformation learner through role-ordered VLM/VAE conditioning and training-only DINOv3 relation alignment. The complementary pathways carry semantic and visual evidence, while relation supervision strengthens transfer across content-distinct scenes without an added inference-time module. The \dataset checkpoint broadens the range of held-out operations addressed by the same interface. Together, the method and corpus support a shared demonstration-based interface for diverse visual transformations.

\paragraph{Broader impacts.}
Exemplar-guided editing can transfer harmful changes as readily as beneficial ones, while synthetic and web-derived examples may carry visual biases.
Evaluation of both transformation transfer and retained query content should accompany deployment.

\clearpage
\bibliographystyle{iclr2027_conference}
\bibliography{references}

\begin{thebibliography}{54}
\providecommand{\natexlab}[1]{#1}
\providecommand{\url}[1]{\texttt{#1}}
\expandafter\ifx\csname urlstyle\endcsname\relax
  \providecommand{\doi}[1]{doi: #1}\else
  \providecommand{\doi}{doi: \begingroup \urlstyle{rm}\Url}\fi

\bibitem[Abdelhamed et~al.(2018)Abdelhamed, Lin, and Brown]{sidd}
A.~Abdelhamed, S.~Lin, and M.~S. Brown.
\newblock A high-quality denoising dataset for smartphone cameras.
\newblock In \emph{Proceedings of the IEEE Conference on Computer Vision and
  Pattern Recognition}, pp.\  1692--1700, 2018.

\bibitem[Abuolaim \& Brown(2020)Abuolaim and Brown]{dpdd}
A.~Abuolaim and M.~S. Brown.
\newblock Defocus deblurring using dual-pixel data.
\newblock In \emph{European Conference on Computer Vision}, pp.\  111--126,
  2020.

\bibitem[Afifi \& Brown(2020)Afifi and Brown]{renderedwb}
M.~Afifi and M.~S. Brown.
\newblock Deep white-balance editing.
\newblock In \emph{Proceedings of the IEEE/CVF Conference on Computer Vision
  and Pattern Recognition}, pp.\  1397--1406, 2020.

\bibitem[Bai et~al.(2025)]{qwenvl}
S.~Bai et~al.
\newblock {Qwen2.5-VL} technical report, 2025.

\bibitem[Bar et~al.(2022)Bar, Gandelsman, Darrell, Globerson, and
  Efros]{visualprompt}
A.~Bar, Y.~Gandelsman, T.~Darrell, A.~Globerson, and A.~A. Efros.
\newblock Visual prompting via image inpainting.
\newblock In \emph{Advances in Neural Information Processing Systems}, 2022.

\bibitem[Brooks et~al.(2023)Brooks, Holynski, and Efros]{instructpix2pix}
T.~Brooks, A.~Holynski, and A.~A. Efros.
\newblock Instructpix2pix: Learning to follow image editing instructions.
\newblock In \emph{Proceedings of the IEEE/CVF Conference on Computer Vision
  and Pattern Recognition}, 2023.

\bibitem[Chen et~al.(2026{\natexlab{a}})Chen, Li, Fu, Zhao, Liu, Liang, Li, and
  Mao]{deltaadapter}
J.~Chen, S.~Li, H.~Fu, B.~Zhao, W.~Liu, Y.~Liang, Q.~Li, and X.~Mao.
\newblock {Delta-Adapter}: Scalable exemplar-based image editing with
  single-pair supervision, 2026{\natexlab{a}}.

\bibitem[Chen et~al.(2025)Chen, Mao, Gu, and Shou]{edittransfer}
L.~Chen, Q.~Mao, Y.~Gu, and M.~Z. Shou.
\newblock Edit transfer: Learning image editing via vision in-context
  relations, 2025.

\bibitem[Chen et~al.(2026{\natexlab{b}})Chen, Mao, Song, Gu, and
  Ma]{edittransferpp}
L.~Chen, Q.~Mao, Y.~Song, Y.~Gu, and S.~Ma.
\newblock {EditTransfer++}: Toward faithful and efficient visual-prompt-guided
  image editing, 2026{\natexlab{b}}.

\bibitem[Chen et~al.(2024)Chen, Huang, Liu, Shen, Zhao, and Zhao]{anydoor}
X.~Chen, L.~Huang, Y.~Liu, Y.~Shen, D.~Zhao, and H.~Zhao.
\newblock {AnyDoor}: Zero-shot object-level image customization.
\newblock In \emph{Proceedings of the IEEE/CVF Conference on Computer Vision
  and Pattern Recognition}, 2024.

\bibitem[Chin(2025)]{openphotorestore}
J.~Chin.
\newblock {OpenPhoto Restore Dataset}.
\newblock Hugging Face dataset, 2025.
\newblock URL
  \url{https://huggingface.co/datasets/joshuachin/openphoto-restore-dataset}.

\bibitem[Couairon et~al.(2023)Couairon, Verbeek, Schwenk, and Cord]{diffedit}
G.~Couairon, J.~Verbeek, H.~Schwenk, and M.~Cord.
\newblock {DiffEdit}: Diffusion-based semantic image editing with mask
  guidance.
\newblock In \emph{International Conference on Learning Representations}, 2023.

\bibitem[Gao et~al.(2026)Gao, Liu, Li, Sun, Tu, Shen, Zhang, Zhao, and
  Zhang]{megastyle}
J.~Gao, S.~Liu, J.~Li, Y.~Sun, Y.~Tu, F.~Shen, W.~Zhang, C.~Zhao, and J.~Zhang.
\newblock {MegaStyle}: Constructing diverse and scalable style dataset via
  consistent text-to-image style mapping, 2026.

\bibitem[Geng et~al.(2024)Geng, Yang, Hang, Li, Gu, Zhang, Bao, Zhang, Hu,
  Chen, and Guo]{instructdiffusion}
Z.~Geng, B.~Yang, T.~Hang, C.~Li, S.~Gu, T.~Zhang, J.~Bao, Z.~Zhang, H.~Hu,
  D.~Chen, and B.~Guo.
\newblock {InstructDiffusion}: A generalist modeling interface for vision
  tasks.
\newblock In \emph{Proceedings of the IEEE/CVF Conference on Computer Vision
  and Pattern Recognition}, 2024.

\bibitem[Gong et~al.(2025)Gong, Song, Li, Li, and Zhang]{relationadapter}
Y.~Gong, Y.~Song, Y.~Li, C.~Li, and Y.~Zhang.
\newblock {RelationAdapter}: Learning and transferring visual relation with
  diffusion transformers, 2025.

\bibitem[Gupta et~al.(2019)Gupta, Doll\'ar, and Girshick]{lvis}
A.~Gupta, P.~Doll\'ar, and R.~Girshick.
\newblock {LVIS}: A dataset for large vocabulary instance segmentation.
\newblock In \emph{Proceedings of the IEEE/CVF Conference on Computer Vision
  and Pattern Recognition}, 2019.

\bibitem[Hertz et~al.(2022)Hertz, Mokady, Tenenbaum, Aberman, Pritch, and
  Cohen-Or]{prompttoprompt}
A.~Hertz, R.~Mokady, J.~Tenenbaum, K.~Aberman, Y.~Pritch, and D.~Cohen-Or.
\newblock Prompt-to-prompt image editing with cross attention control, 2022.

\bibitem[Hertzmann et~al.(2001)Hertzmann, Jacobs, Oliver, Curless, and
  Salesin]{imageanalogies}
A.~Hertzmann, C.~E. Jacobs, N.~Oliver, B.~Curless, and D.~H. Salesin.
\newblock Image analogies.
\newblock In \emph{Proceedings of ACM SIGGRAPH}, 2001.

\bibitem[Hu et~al.(2022)]{lora}
E.~Hu et~al.
\newblock {LoRA}: Low-rank adaptation of large language models.
\newblock In \emph{International Conference on Learning Representations}, 2022.

\bibitem[Huang et~al.(2024)Huang, Xie, Wang, Yuan, Cun, Ge, Zhou, Dong, Huang,
  Zhang, and Shan]{smartedit}
Y.~Huang, L.~Xie, X.~Wang, Z.~Yuan, X.~Cun, Y.~Ge, J.~Zhou, C.~Dong, R.~Huang,
  R.~Zhang, and Y.~Shan.
\newblock {SmartEdit}: Exploring complex instruction-based image editing with
  multimodal large language models.
\newblock In \emph{Proceedings of the IEEE/CVF Conference on Computer Vision
  and Pattern Recognition}, 2024.

\bibitem[Li et~al.(2019)Li, Ren, Fu, Tao, Feng, Zeng, and Wang]{reside}
B.~Li, W.~Ren, D.~Fu, D.~Tao, D.~Feng, W.~Zeng, and Z.~Wang.
\newblock Benchmarking single-image dehazing and beyond.
\newblock \emph{IEEE Transactions on Image Processing}, 28\penalty0
  (1):\penalty0 492--505, 2019.

\bibitem[Li et~al.(2020{\natexlab{a}})Li, Guo, Ren, Cong, Hou, Kwong, and
  Tao]{uieb}
C.~Li, C.~Guo, W.~Ren, R.~Cong, J.~Hou, S.~Kwong, and D.~Tao.
\newblock An underwater image enhancement benchmark dataset and beyond.
\newblock \emph{IEEE Transactions on Image Processing}, 29:\penalty0
  4376--4389, 2020{\natexlab{a}}.

\bibitem[Li et~al.(2022)Li, Zhang, Zhang, Huang, Tian, and Tao]{realrain1k}
W.~Li, Q.~Zhang, J.~Zhang, Z.~Huang, X.~Tian, and D.~Tao.
\newblock Toward real-world single image deraining: A new benchmark and beyond,
  2022.

\bibitem[Li et~al.(2020{\natexlab{b}})Li, Wei, Chen, Tai, and Tang]{fss1000}
X.~Li, T.~Wei, Y.~P. Chen, Y.-W. Tai, and C.-K. Tang.
\newblock {FSS-1000}: A 1000-class dataset for few-shot segmentation.
\newblock In \emph{Proceedings of the IEEE/CVF Conference on Computer Vision
  and Pattern Recognition}, 2020{\natexlab{b}}.

\bibitem[Li et~al.(2026)Li, Duan, Ye, Chen, Chen, Li, and Chen]{viral}
Z.~Li, Z.~Duan, J.~Ye, C.~Chen, D.~Chen, Y.~Li, and Y.~Chen.
\newblock {VIRAL}: Visual in-context reasoning via analogy in diffusion
  transformers, 2026.

\bibitem[Lim et~al.(2017)Lim, Son, Kim, Nah, and Lee]{flickr2k}
B.~Lim, S.~Son, H.~Kim, S.~Nah, and K.~M. Lee.
\newblock Enhanced deep residual networks for single image super-resolution.
\newblock In \emph{Proceedings of the IEEE Conference on Computer Vision and
  Pattern Recognition Workshops}, pp.\  136--144, 2017.

\bibitem[Lipman et~al.(2023)Lipman, Chen, Ben-Hamu, Nickel, and
  Le]{flowmatching}
Y.~Lipman, R.~T.~Q. Chen, H.~Ben-Hamu, M.~Nickel, and M.~Le.
\newblock Flow matching for generative modeling.
\newblock In \emph{International Conference on Learning Representations}, 2023.

\bibitem[Liu et~al.(2023)Liu, Gong, and Liu]{rectifiedflow}
X.~Liu, C.~Gong, and Q.~Liu.
\newblock Flow straight and fast: Learning to generate and transfer data with
  rectified flow.
\newblock In \emph{International Conference on Learning Representations}, 2023.

\bibitem[Liu et~al.(2018)Liu, Jaw, Huang, and Hwang]{snow100k}
Y.-F. Liu, D.-W. Jaw, S.-C. Huang, and J.-N. Hwang.
\newblock {DesnowNet}: Context-aware deep network for snow removal.
\newblock \emph{IEEE Transactions on Image Processing}, 27\penalty0
  (6):\penalty0 3064--3073, 2018.

\bibitem[Lu et~al.(2025)Lu, Chen, Yang, Gnanha, Wang, Li, and Mao]{pairedit}
H.~Lu, J.~Chen, Z.~Yang, A.~T. Gnanha, F.~L. Wang, Q.~Li, and X.~Mao.
\newblock {PairEdit}: Learning semantic variations for exemplar-based image
  editing.
\newblock In \emph{Advances in Neural Information Processing Systems}, 2025.

\bibitem[Manor et~al.(2026)Manor, Gal, Maron, Michaeli, and Chechik]{lorweb}
H.~Manor, R.~Gal, H.~Maron, T.~Michaeli, and G.~Chechik.
\newblock Spanning the visual analogy space with a weight basis of {LoRA}s,
  2026.

\bibitem[Nah et~al.(2017)Nah, Kim, and Lee]{gopro}
S.~Nah, T.~H. Kim, and K.~M. Lee.
\newblock Deep multi-scale convolutional neural network for dynamic scene
  deblurring.
\newblock In \emph{Proceedings of the IEEE Conference on Computer Vision and
  Pattern Recognition}, pp.\  3883--3891, 2017.

\bibitem[Qian et~al.(2018)Qian, Tan, Yang, Su, and Liu]{deraindrop}
R.~Qian, R.~T. Tan, W.~Yang, J.~Su, and J.~Liu.
\newblock Attentive generative adversarial network for raindrop removal from a
  single image.
\newblock In \emph{Proceedings of the IEEE Conference on Computer Vision and
  Pattern Recognition}, pp.\  2482--2491, 2018.

\bibitem[Qu et~al.(2017)Qu, Tian, He, Tang, and Lau]{srd}
L.~Qu, J.~Tian, S.~He, Y.~Tang, and R.~W.~H. Lau.
\newblock {DeshadowNet}: A multi-context embedding deep network for shadow
  removal.
\newblock In \emph{Proceedings of the IEEE Conference on Computer Vision and
  Pattern Recognition}, pp.\  4067--4075, 2017.

\bibitem[Radford et~al.(2021)]{clip}
A.~Radford et~al.
\newblock Learning transferable visual models from natural language
  supervision.
\newblock In \emph{International Conference on Machine Learning}, 2021.

\bibitem[Ramanathan et~al.(2023)Ramanathan, Kalia, Petrovic, Wen, Zheng, Guo,
  Wang, Marquez, Kovvuri, Kadian, Mousavi, Song, Dubey, and Mahajan]{paco}
V.~Ramanathan, A.~Kalia, V.~Petrovic, Y.~Wen, B.~Zheng, B.~Guo, R.~Wang,
  A.~Marquez, R.~Kovvuri, A.~Kadian, A.~Mousavi, Y.~Song, A.~Dubey, and
  D.~Mahajan.
\newblock {PACO}: Parts and attributes of common objects.
\newblock In \emph{Proceedings of the IEEE/CVF Conference on Computer Vision
  and Pattern Recognition}, 2023.

\bibitem[Rim et~al.(2020)Rim, Lee, Won, and Cho]{realblur}
J.~Rim, H.~Lee, J.~Won, and S.~Cho.
\newblock Real-world blur dataset for learning and benchmarking deblurring
  algorithms.
\newblock In \emph{European Conference on Computer Vision}, pp.\  184--201,
  2020.

\bibitem[Shafaei et~al.(2021)Shafaei, Little, and Schmidt]{ffhqr}
A.~Shafaei, J.~J. Little, and M.~Schmidt.
\newblock {AutoRetouch}: Automatic professional face retouching.
\newblock In \emph{Proceedings of the IEEE/CVF Winter Conference on
  Applications of Computer Vision}, 2021.

\bibitem[Shen et~al.(2025)Shen, Jiang, He, Ye, Wang, Du, Li, and Tang]{igpair}
F.~Shen, X.~Jiang, X.~He, H.~Ye, C.~Wang, X.~Du, Z.~Li, and J.~Tang.
\newblock {IMAGDressing-v1}: Customizable virtual dressing.
\newblock In \emph{Proceedings of the AAAI Conference on Artificial
  Intelligence}, 2025.

\bibitem[Sim\'eoni et~al.(2025)]{dinov3}
O.~Sim\'eoni et~al.
\newblock {DINOv3}, 2025.

\bibitem[Sun et~al.(2023)Sun, Yang, Peng, Shen, Yang, Hu, Qiu, and
  Koike]{imagebrush}
Y.~Sun, Y.~Yang, H.~Peng, Y.~Shen, Y.~Yang, H.~Hu, L.~Qiu, and H.~Koike.
\newblock {ImageBrush}: Learning visual in-context instructions for
  exemplar-based image manipulation.
\newblock In \emph{Advances in Neural Information Processing Systems}, 2023.

\bibitem[Wan et~al.(2017)Wan, Shi, Duan, Tan, and Kot]{sir2}
R.~Wan, B.~Shi, L.-Y. Duan, A.-H. Tan, and A.~C. Kot.
\newblock Benchmarking single-image reflection removal algorithms.
\newblock In \emph{Proceedings of the IEEE International Conference on Computer
  Vision}, pp.\  3922--3930, 2017.

\bibitem[Wang et~al.(2025{\natexlab{a}})Wang, Cvejic, Eldesokey, and
  Wonka]{editclip}
Q.~Wang, A.~Cvejic, A.~Eldesokey, and P.~Wonka.
\newblock Editclip: Representation learning for image editing.
\newblock In \emph{Proceedings of the IEEE/CVF International Conference on
  Computer Vision}, pp.\  15960--15970, 2025{\natexlab{a}}.

\bibitem[Wang et~al.(2023)]{painter}
X.~Wang et~al.
\newblock Images speak in images: A generalist painter for in-context visual
  learning.
\newblock In \emph{Proceedings of the IEEE/CVF Conference on Computer Vision
  and Pattern Recognition}, 2023.

\bibitem[Wang et~al.(2022)Wang, Wan, Yang, Li, Chau, and Kot]{lolv2}
Y.~Wang, R.~Wan, W.~Yang, H.~Li, L.-P. Chau, and A.~Kot.
\newblock Low-light image enhancement with normalizing flow.
\newblock In \emph{Proceedings of the AAAI Conference on Artificial
  Intelligence}, volume~36, pp.\  2604--2612, 2022.

\bibitem[Wang et~al.(2025{\natexlab{b}})Wang, Liu, Lin, Liu, Yi, Wang, and
  Ma]{omnistyle}
Y.~Wang, R.~Liu, J.~Lin, F.~Liu, Z.~Yi, Y.~Wang, and R.~Ma.
\newblock {OmniStyle}: Filtering high quality style transfer data at scale.
\newblock In \emph{Proceedings of the IEEE/CVF Conference on Computer Vision
  and Pattern Recognition}, pp.\  7847--7856, 2025{\natexlab{b}}.

\bibitem[Wu et~al.(2025)Wu, Bai, et~al.]{qwenimage}
F.~Wu, S.~Bai, et~al.
\newblock {Qwen-Image} technical report, 2025.

\bibitem[Xiao et~al.(2026)Xiao, Zhang, Song, Chen, Li, Jiang, Ren, Lin, Huang,
  Huang, Li, Duan, and Qi]{spatialedit}
Y.~Xiao, W.~Zhang, L.~Song, Y.~Chen, W.~Li, N.~Jiang, T.~Ren, H.~Lin, W.~Huang,
  H.~Huang, X.~Li, N.~Duan, and X.~Qi.
\newblock {SpatialEdit}: Benchmarking fine-grained image spatial editing, 2026.

\bibitem[Xu et~al.(2025)Xu, Fan, Kou, Qin, Gu, Zhao, Ling, and
  Wang]{textualizevisualprompt}
P.~Xu, Q.~Fan, F.~Kou, S.~Qin, H.~Gu, R.~Zhao, C.~Ling, and B.~Wang.
\newblock Textualize visual prompt for image editing via diffusion bridge.
\newblock In \emph{Proceedings of the AAAI Conference on Artificial
  Intelligence}, volume~39, pp.\  21779--21787, 2025.
\newblock \doi{10.1609/aaai.v39i20.35483}.

\bibitem[Yang et~al.(2023)Yang, Gu, Zhang, Zhang, Chen, Sun, Chen, and
  Wen]{paintbyexample}
B.~Yang, S.~Gu, B.~Zhang, T.~Zhang, X.~Chen, X.~Sun, D.~Chen, and F.~Wen.
\newblock Paint by example: Exemplar-based image editing with diffusion models.
\newblock In \emph{Proceedings of the IEEE/CVF Conference on Computer Vision
  and Pattern Recognition}, 2023.

\bibitem[Ye et~al.(2023)Ye, Zhang, Liu, Han, and Yang]{ipadapter}
H.~Ye, J.~Zhang, S.~Liu, X.~Han, and W.~Yang.
\newblock {IP-Adapter}: Text compatible image prompt adapter for text-to-image
  diffusion models, 2023.

\bibitem[Yu et~al.(2022)Yu, Dai, Li, Ma, Shen, Li, and Qi]{uhdm}
X.~Yu, P.~Dai, W.~Li, L.~Ma, J.~Shen, J.~Li, and X.~Qi.
\newblock Towards efficient and scale-robust ultra-high-definition image
  demoir{\'e}ing.
\newblock In \emph{European Conference on Computer Vision}, 2022.

\bibitem[Zhang et~al.(2023)Zhang, Rao, and Agrawala]{controlnet}
L.~Zhang, A.~Rao, and M.~Agrawala.
\newblock Adding conditional control to text-to-image diffusion models.
\newblock In \emph{Proceedings of the IEEE/CVF International Conference on
  Computer Vision}, 2023.

\bibitem[Zhang et~al.(2018)Zhang, Isola, Efros, Shechtman, and Wang]{lpips}
R.~Zhang, P.~Isola, A.~A. Efros, E.~Shechtman, and O.~Wang.
\newblock The unreasonable effectiveness of deep features as a perceptual
  metric.
\newblock In \emph{Proceedings of the IEEE/CVF Conference on Computer Vision
  and Pattern Recognition}, 2018.

\end{thebibliography}
\appendix
\clearpage
\raggedbottom

\section{Evaluation Scope}
\label{app:evidence}

\paragraph{Comparison design.}
One-pair transfer is established in prior work; CrossPair's contribution is its role-ordered use of the editor's native conditioning paths and training-only DINO relation alignment.
We compare released systems as complete pipelines and use the matched ablation to examine CrossPair's components.
The Relation252K-derived model is the core method comparison; the Transform-360K checkpoint assesses the combined effect of its larger, differently composed training corpus.

\paragraph{Metric interpretation.}
CLIP and LPIPS results are query-level micro averages across task families, while GPT-5.5 provides one ranking per query.
Source similarity and the content ranking complement target agreement by revealing the transformation--preservation trade-off in Table~\ref{tab:relation252k-main}.
These aggregate measures characterize the sampled benchmark; operation-specific accuracy and pixel-level preservation require task-specific annotations.

\section{\dataset Construction and Datasheet}
\label{app:dataset}

\subsection{Training-unit schema}

Let \(T_g\) denote the transformation associated with relation key \(g\).
A record contains a demonstration pair \((\rs,\rt)\), a query pair
\((\qs,\qt)\), and a natural-language instruction.  The two pairs are selected
from the same construction group under the content-disjoint protocol defined in
the main paper.  The common assembler hard-enforces that they are different
records, so the loader receives one non-self demonstration for every query;
source-specific builders provide the stricter identity/scene predicates.
The serialized prompt is not uniformly role-only: Relation252K records, for
example, contain the role instruction followed by the concrete edit
description.  Exactly one demonstration pair is used in the frozen
\dataset manifest and at inference.

\begin{center}
\centering
\small
\setlength{\tabcolsep}{4pt}
\begin{tabular}{@{}p{0.40\columnwidth}p{0.50\columnwidth}@{}}
\toprule
Field & Meaning \\
\midrule
{\urlstyle{tt}\nolinkurl{edit_pairs[0].}}\newline\texttt{before} &
demonstration before image \(\rs\) \\
{\urlstyle{tt}\nolinkurl{edit_pairs[0].}}\newline\texttt{after} &
demonstration after image \(\rt\) \\
{\urlstyle{tt}\nolinkurl{source_image}}         & query source image \(\qs\) \\
\texttt{image}                 & query target image \(\qt\), training/evaluation only \\
\texttt{prompt}                & role instruction, optionally followed by an edit description \\
\texttt{style}                 & serialized construction/group label \\
\texttt{task}                  & optional task-group label; used by the Relation252K segment \\
{\urlstyle{tt}\nolinkurl{source_dataset}}       & optional source-family provenance \\
\texttt{resize} & optional top-level longest-side resize metadata \\
{\urlstyle{tt}\nolinkurl{edit_pairs[0].}}\newline\texttt{resize} & optional pair-level longest-side resize metadata \\
\bottomrule
\end{tabular}
\captionsetup{type=table,hypcap=false}\caption{Fields consumed by the frozen \dataset loader.  Additional
provenance fields occur in source-specific segments.}
\label{tab:record-schema}
\end{center}

\subsection{Frozen manifest statistics}

The frozen training metadata is
{\urlstyle{tt}\nolinkurl{train_final_with_megastyle_revctl.json}}.  A full manifest scan
contains 91,010 records.  Because every record has \(K=1\), it references four
image slots: query source, query target, demonstration before, and demonstration
after.  Thus the corpus name refers to 364,040 record-level image slots, not to
unique files.  The same scan finds 248,910 unique path strings and 15,428
distinct prompt strings.  It also finds 46,819 distinct non-Relation252K
\texttt{style} values; the Relation252K segment uses one shared
\texttt{style} value and preserves 220 task groups in \texttt{task}.  These
serialized labels are an implementation inventory, not a replacement for the
paper-level count of over 8K fine-grained relations: several labels are
source-local identifiers and are not counted as separately audited relation
types in the paper.

\begin{table*}[tp]
\caption{Composition of the frozen \dataset manifest.  Counts in the fourth
column are serialized labels, not the paper-level relation count; segment-local
labels can overlap.}
\label{tab:construction-families}
\centering
\small
\setlength{\tabcolsep}{4pt}
\begin{tabularx}{\textwidth}{@{}Xrrrrr@{}}
\toprule
Manifest segment & Records & Share & Styles & Slots & Paths \\
\midrule
Low-level vision & 20,049 & 22.03\% & 20,049 & 80,196 & 80,196 \\
Relation252K & 17,600 & 19.34\% & 1 (220 \texttt{task}) & 70,400 & 16,994 \\
Virtual try-on & 13,527 & 14.86\% & 3,936 & 54,108 & 51,436 \\
Conditional representations & 11,645 & 12.80\% & 196 & 46,580 & 14,402 \\
Segmentation/reference perception & 10,443 & 11.47\% & 9,723 & 41,772 & 18,387 \\
MegaStyle & 7,552 & 8.30\% & 7,552 & 30,208 & 30,208 \\
Civitai style & 4,802 & 5.28\% & 3,450 & 19,208 & 17,658 \\
OmniStyle & 2,685 & 2.95\% & 896 & 10,740 & 8,950 \\
Spatial transformations & 2,144 & 2.36\% & 456 & 8,576 & 8,450 \\
Other paired sources & 563 & 0.62\% & 563 & 2,252 & 2,252 \\
\midrule
\textbf{Total} & \textbf{91,010} & \textbf{100\%} & \textbf{46,819 global values} & \textbf{364,040} & \textbf{248,910} \\
\bottomrule
\end{tabularx}

\end{table*}

The ten manifest rows roll up to the six construction streams named in the main
paper: the four style-derived rows belong to style transfer, while the imported
Relation252K segment is listed separately for provenance rather than counted as
an additional construction pipeline.

\subsection{Source-specific pair construction}

\paragraph{Style transfer.}
MegaStyle~\citep{megastyle} supplies stylized targets and style metadata.
Qwen-Image-Edit-2511 is used in the retained construction pipeline to generate
a photorealistic counterpart when a paired source is unavailable.  Progressive
subsampling produces 12,450 one-pair style records, of which 7,552 pass the
pair-level semantic audit described below.  OmniStyle~\citep{omnistyle}
contributes 2,685 final records.  Civitai-derived and other style segments
contribute 4,802 and 563 records, respectively.  Their serialized prompts name
the style/group identity rather than replacing the visual demonstration.

\paragraph{Conditional representations and segmentation.}
The construction script applies {\urlstyle{tt}\nolinkurl{controlnet_aux}} processors for Canny,
TEED/soft edges, Depth-Anything-3 depth, Marigold normals, DWPose, MLSD,
scribbles, anime/realistic/coarse/standard line art, and shuffled image
conditions.  It retains RGB-to-condition, condition-to-RGB, and selected
condition-to-condition directions as distinct records.  The reversed segment
explicitly stores {\urlstyle{tt}\nolinkurl{condition_type}}, {\urlstyle{tt}\nolinkurl{original_task}}, and
\texttt{direction}; cross-condition pairs include depth--edge,
normal--edge, and depth--normal.  Segmentation/reference-perception records are
drawn from the checked source manifests for COCO, LVIS~\citep{lvis},
FSS-1000~\citep{fss1000}, PACO~\citep{paco}, Pascal-Part, and Cityscapes.

\paragraph{Virtual try-on.}
The try-on pipeline groups IGPair\slash IMAGDressing-v1 examples~\citep{igpair} by
garment identifier, constructs person--garment inputs, and uses the retained
FireRed inference outputs as paired targets.  The final manifest contains
13,527 records over 3,936 garment labels.

\paragraph{Spatial transformations.}
The SpatialEdit-derived segment~\citep{spatialedit} contains 2,144 records over
456 serialized labels.  Labels encode signed yaw and pitch values and a
distance state, for example
{\urlstyle{tt}\nolinkurl{y+000_p+000_dneg__armchair}}, across nine object categories.

\paragraph{Low-level vision.}
The pre-filter low-level manifest contains 21,078 groups; 20,049 records remain
in the frozen training manifest.  Table~\ref{tab:lowlevel-inventory} reports the
group labels and counts read from that manifest rather than inferring sources
that are absent from it.

\begin{table*}[tp]
\caption{Low-level group counts in the checked construction manifest.}
\label{tab:lowlevel-inventory}
\centering
\small
\setlength{\tabcolsep}{4pt}
\begin{tabular}{@{}p{0.23\textwidth}p{0.69\textwidth}@{}}
\toprule
Family & Manifest groups before final mixture assembly \\
\midrule
Denoising & SIDD~\citep{sidd} (80) \\
Deblurring & GoPro~\citep{gopro} (63), RealBlur~\citep{realblur} (234), DPDD~\citep{dpdd} (250) \\
Resolution / moir\'e & Flickr2K~\citep{flickr2k} (1,325), UHDM~\citep{uhdm} (1,260) \\
Weather removal & \texttt{haze} (746), \texttt{rain} (392), \texttt{snow} (750), \texttt{raindrop} (505)~\citep{reside,realrain1k,snow100k,deraindrop} \\
Illumination / color & LOLv2~\citep{lolv2} (894), six Rendered-WB subsets~\citep{renderedwb} (853 each), UIEB~\citep{uieb} (346) \\
Reflection / shadow & SIR2~\citep{sir2} (72), SRD~\citep{srd} (1,543) \\
Portrait / photo restoration & {\urlstyle{tt}\nolinkurl{ffhqr_out}} (5,000), {\urlstyle{tt}\nolinkurl{photo_restore}} (2,500)~\citep{ffhqr,openphotorestore} \\
\bottomrule
\end{tabular}

\end{table*}

\paragraph{Imported Relation252K pairs.}
The controlled Relation252K source contributes 17,600 records.  Its serialized
\texttt{style} is shared, while \texttt{task} preserves 220 task/group labels;
for example, {\urlstyle{tt}\nolinkurl{1-TeethAlignment/Group1}}.  The prompt contains both the
three-image role specification and the concrete edit description.

\subsection{Quality control}

Every image is fully decoded, non-positive dimensions are rejected, and the
same aspect-preserving, multiple-of-16 geometry preprocessing used in training
is executed as a preflight check.  Auxiliary loaders reject images with a side
shorter than eight pixels and reject truncated files by default.  A pass over
72,273 captioned records retains 67,271 and removes 5,002 decoding or geometry
failures.  Segmentation masks covering less than 10\% of an image are removed,
and a manually curated style bad-list removes 281 folders or indices.  The
MegaStyle semantic audit evaluates 12,450 pairs and retains a pair only when
both images receive \texttt{YES}; 7,552 pairs pass (60.7\%).  These checks are
combined with the source-specific automatic and semantic validation described
in the main paper.

\subsection{Grouping and record formation}

The common assembler groups source pairs by the operation key, subsamples and
caps source-specific groups, and selects a non-self peer as the demonstration.
Source-specific builders enforce the distinct-content/scene constraint stated
in the main paper; low-level records additionally forbid references with the
same internal {\urlstyle{tt}\nolinkurl{_group_id}}.  Assembly uses random seed 42.  Intermediate
builders support \(K\in\{1,2,3\}\), whereas the frozen \dataset manifest is
entirely \(K=1\).  Thus 91,010 records yield 364,040 record-level image slots.

\subsection{Evaluation sets}

\begin{center}
\centering
\small
\setlength{\tabcolsep}{4pt}
\begin{tabular}{@{}p{0.76\columnwidth}r@{}}
\toprule
Metadata collection & Records \\
\midrule
Relation252K seen-task held-out set (219 groups \(\times\) 5) & 1,095 \\
Released unseen-task sample (30 categories) & 300 \\
Additional OOD candidates in the paper protocol & 150 \\
\bottomrule
\end{tabular}
\captionsetup{type=table,hypcap=false}\caption{Evaluation-set sizes used by the paper protocol.}
\label{tab:eval-splits}
\end{center}

The seen-task builder reserves the last five eligible images per task after
excluding tasks with at most 19 images.  The OOD protocol combines 300 released
unseen cases with 150 additional cases from diverse transformation families.
Non-self sampling prevents a query record from using itself as its
demonstration, and held-out queries are excluded from the corresponding
training groups.

\section{Training Details}
\label{app:training}

\subsection{Architecture and trainable parameters}

\method is instantiated from Qwen-Image-Edit-2511~\citep{qwenimage}.  The
demonstration before, demonstration after, and query source are serialized in
that order in both native conditioning pathways.  The VLM, VAE, base image
generator, and DINOv3 encoder~\citep{dinov3} are frozen.  Only rank-128
LoRA adapters~\citep{lora} attached to the image generator are optimized.

The checked implementation targets
{\urlstyle{tt}\nolinkurl{to_q}}, {\urlstyle{tt}\nolinkurl{to_k}}, {\urlstyle{tt}\nolinkurl{to_v}},
{\urlstyle{tt}\nolinkurl{add_q_proj}}, {\urlstyle{tt}\nolinkurl{add_k_proj}},
{\urlstyle{tt}\nolinkurl{add_v_proj}}, {\urlstyle{tt}\nolinkurl{to_out.0}},
{\urlstyle{tt}\nolinkurl{to_add_out}}, {\urlstyle{tt}\nolinkurl{img_mlp.net.2}},
{\urlstyle{tt}\nolinkurl{img_mod.1}}, {\urlstyle{tt}\nolinkurl{txt_mlp.net.2}}, and
{\urlstyle{tt}\nolinkurl{txt_mod.1}}.

\subsection{Optimization configuration}

Table~\ref{tab:training-config} lists the training configuration.

\begin{table*}[tp]
\caption{Training configuration.}
\label{tab:training-config}
\centering
\small
\setlength{\tabcolsep}{5pt}
\begin{tabular}{@{}p{0.34\textwidth}p{0.58\textwidth}@{}}
\toprule
Item & Setting \\
\midrule
Optimizer & AdamW, weight decay 0.01, \(\beta=(0.9,0.999)\), \(\epsilon=10^{-8}\) \\
Learning rate / schedule & \(10^{-4}\) / \texttt{ConstantLR} \\
LoRA rank / target & 128 / modules listed above \\
Per-device batch / accumulation & 1 record / 1 step \\
Precision / distributed optimizer & bfloat16 / DeepSpeed ZeRO-2, no CPU offload \\
Gradient checkpointing & enabled \\
Unused-parameter detection & enabled \\
Base/data-order seed & 20260719; each rank adds its process index \\
Other auxiliary losses & disabled \\
\bottomrule
\end{tabular}

\end{table*}

For each record, the training step draws one scheduler index uniformly from
the configured timestep interval (the checked default spans the full interval),
draws Gaussian noise with scale 1, and forms the native Qwen flow-matching
target.  The primary loss is MSE multiplied by the scheduler's training weight.
Training uses {\urlstyle{tt}\nolinkurl{cfg_scale=1}}.  No warmup or decay is introduced by the
\texttt{ConstantLR} scheduler.

\paragraph{Distributed sampling.}
The K-bucket sampler groups records by the number of demonstration pairs so all
ranks execute compatible tensor shapes.  With \(K=1\), every rank receives one
record per optimizer step.  Corrupt images are not retried indefinitely:
the checked loader attempts a path once and then tries at most eight replacement
records from the same \(K\) bucket, logging failures as JSONL.

\subsection{Resolution and image preprocessing}

The frozen \dataset configuration uses an area cap of 262,144 pixels and
512-pixel longest-side metadata.  The loader preserves aspect ratio, rounds
dimensions down to multiples of 16, and applies bilinear longest-side resize.
No random crop, flip, or color jitter appears in the checked training path.
The VLM wrapper independently resizes references toward a \(384^2\)-pixel area,
whereas the VAE reference branch uses its native \(1024^2\)-pixel area target.
DINOv3 receives a separate bicubic 224-by-224 input.

\subsection{Role prompt and VLM serialization}
\label{app:prompt-evidence}

The training code uses \texttt{data["prompt"]}, retaining record-specific
instructions. A Relation252K example begins with:
\begin{quote}
\small\ttfamily\raggedright
Edit image 3 based on the transformation from image 1 to image 2.
Align the teeth and enhance teeth whiteness
\end{quote}
The VAE-only ablation instead uses the exact fallback
{\urlstyle{tt}\nolinkurl{Edit the input image.}}, because its VLM branch receives only the query
source.  The image order is always demonstration before, demonstration after,
then query source for \(K=1\).

The Qwen VLM wrapper~\citep{qwenvl} uses the following fixed system instruction and
role-ordered image serialization.  The three image placeholders correspond to
\(\rs,\rt,\qs\), respectively.

\begin{figure*}[t]
\begin{minipage}{0.97\textwidth}
\small
\textbf{System instruction}
\begin{verbatim}
Describe the key features of the input image (color, shape, size,
texture, objects, background), then explain how the user's text
instruction should alter or modify the image. Generate a new image
that meets the user's requirements while maintaining consistency
with the original input where appropriate.
\end{verbatim}
\textbf{User serialization}
\begin{verbatim}
Picture 1: <|vision_start|><|image_pad|><|vision_end|>
Picture 2: <|vision_start|><|image_pad|><|vision_end|>
Picture 3: <|vision_start|><|image_pad|><|vision_end|>
<record-specific prompt>
\end{verbatim}
\end{minipage}
\caption{Exact semantic content of the VLM prompt template in the inspected
implementation.  Whitespace is shown for readability; the code concatenates
the three image placeholders before the record-specific prompt.}
\label{fig:vlm-prompt}
\end{figure*}

\subsection{DINOv3 implementation details}

This section records implementation details not repeated with the derivation in
the main paper.  The frozen encoder is DINOv3 ViT-7B/16.  Inputs are resized to
224-by-224 and normalized with mean \((0.485,0.456,0.406)\) and standard
deviation \((0.229,0.224,0.225)\).  The implementation applies the final model
normalization, removes the CLS token and the configured number of register
tokens, and retains the \(14\times14=196\) patch tokens.  All encoder parameters
are frozen; only the predicted-image branch preserves its input gradient.

For the DINO relation equations in the main paper, change scores are detached, clamped
non-negative, raised to \(\alpha=1\), and normalized over patches.  The query
map in {\urlstyle{tt}\nolinkurl{source_gt}} mode is computed from detached query-source and target
features; it localizes the edit but never replaces the differentiable
source-to-prediction direction.  No cross-scene patch correspondence is used.

The relation configuration uses
{\urlstyle{tt}\nolinkurl{dino_mode=pair_relation}}, \(\lambda_{\mathrm{DINO}}=0.003\),
\(\sigma_{\max}=0.5\), linear \(w_\sigma\), {\urlstyle{tt}\nolinkurl{dino_every=1}},
\(\alpha=1\), {\urlstyle{tt}\nolinkurl{source_gt}} query weights, 224-pixel DINO input, and a
gradient probe every 50 steps.  Direct perceptual, delta-consistency, and QwenVL
relation losses are disabled.

\section{Inference Details}
\label{app:inference}

Inference routes \((\rs,\rt,\qs)\) through both the VLM and VAE branches in the
fixed order used during training.  DINOv3, the target image, and all auxiliary
loss code are removed.  Table~\ref{tab:inference-config} records the settings
used by the reported CrossPair launcher.

\begin{center}
\centering
\small
\setlength{\tabcolsep}{4pt}
\begin{tabular}{@{}p{0.42\columnwidth}p{0.48\columnwidth}@{}}
\toprule
Item & Setting \\
\midrule
Sampling steps & 40 \\
Guidance scale & 4.0 \\
Negative prompt & empty string \\
Denoising strength & 1.0 \\
Scheduler & native Qwen-Image scheduler \\
Initial seed & 42 \\
Per-case seed & \(42+\) enumeration index \\
Output size & query-source width and height \\
VAE/VLM reference resize & enabled; native \(1024^2\)/\(384^2\) area targets \\
{\urlstyle{tt}\nolinkurl{zero_cond_t}} & enabled for Qwen-Image-Edit-2511 \\
Conditioning route & both VLM and VAE \\
DINOv3 at inference & absent \\
Precision / batch & bfloat16 / one query per process \\
\bottomrule
\end{tabular}
\captionsetup{type=table,hypcap=false}\caption{CrossPair inference settings.  In the reported sequential
non-category runs, the enumeration index equals the original metadata index.}
\label{tab:inference-config}
\end{center}

\begin{table*}[!t]
\caption{Checked baseline inference interfaces.  Each method retains its
released sampler; all outputs share the same metric preprocessing and
evaluation resolution.}
\label{tab:baseline-protocol}
\centering
\small
\setlength{\tabcolsep}{3pt}
\begin{tabular}{@{}p{0.18\textwidth}p{0.31\textwidth}p{0.09\textwidth}p{0.10\textwidth}p{0.22\textwidth}@{}}
\toprule
Method & Visual input interface & Steps & Guidance & Seed / output handling \\
\midrule
Qwen Edit 2511 & one pair in the VLM route; query source remains in both native routes & 40 & 4.0 & \(42+i\); native output size \\
LoRWeB & one pair and query assembled as a \(2\times2\) grid; lower-right output crop & repo. default & 2.5 & 50; crop restored to source size \\
RelationAdapter & two conditioning images plus query-source VAE input & 24 & 3.5 & 1000; native output \\
CrossPair & one pair and query through both VLM and VAE routes & 40 & 4.0 & \(42+i\); native output size \\
\bottomrule
\end{tabular}

\end{table*}

\begin{figure*}[!t]
\centering
\begin{minipage}[t]{0.48\textwidth}
\small\textbf{System prompt (exact text)}
\begin{Verbatim}[fontsize=\footnotesize,breaklines=true,breakanywhere=true,breaksymbolleft={},breaksymbolright={},baselinestretch=1]
You are a strict visual judge for reference-pair-guided image editing.
You compare K anonymized candidate outputs for the same query.

Judge only from the supplied images. Never infer method identity from
rendering style, candidate position, filenames, or prior knowledge.
Candidate order is randomized.

First infer the intended transformation from REFERENCE BEFORE to
REFERENCE AFTER. Then compare QUERY SOURCE with each candidate.

Use these criteria:
1. Transformation faithfulness: how accurately the demonstrated
   transformation is transferred.
2. Query-content preservation: identity, geometry, layout, pose,
   background, object count, and non-target content should remain
   unchanged unless the demonstration requires otherwise.
3. Visual quality: realism, sharpness, coherence, and absence of
   artifacts.

Return complete rankings without ties. For the overall ranking,
prioritize transformation faithfulness, then content preservation,
then visual quality. Return JSON only.
\end{Verbatim}
\end{minipage}\hfill
\begin{minipage}[t]{0.48\textwidth}
\small\textbf{User prompt (exact template; candidate lines expand to \(1,\ldots,K\))}
\begin{Verbatim}[fontsize=\footnotesize,breaklines=true,breakanywhere=true,breaksymbolleft={},breaksymbolright={},baselinestretch=1]
Images are provided in this order:
REFERENCE BEFORE
REFERENCE AFTER
QUERY SOURCE
CANDIDATE 1
...
CANDIDATE K

Return exactly this JSON schema:
{
  "inferred_transformation": "one concise sentence",
  "overall_ranking": ["candidate_1", ..., "candidate_K"],
  "transformation_ranking": ["candidate_1", ..., "candidate_K"],
  "content_ranking": ["candidate_1", ..., "candidate_K"],
  "rank_1_reason": "one concise sentence",
  "rank_K_reason": "one concise sentence"
}

Each ranking must contain every candidate exactly once.
Do not output markdown.
\end{Verbatim}
\end{minipage}
\caption{Complete GPT-5.5 judge prompt.  In the actual system prompt,
\texttt{K} is replaced by 4 or 5; the JSON arrays are expanded to contain all
candidate identifiers exactly once.}
\label{fig:gpt-prompt}
\end{figure*}

The script uses the native Qwen flow-matching scheduler.  Apart from seeded
initial noise, no extra stochastic operation is introduced by the wrapper.
Deterministic CUDA algorithms are not explicitly enabled.  Generated PNG files
are saved without an additional enhancement, color conversion, or recompression
stage before metric evaluation.

\section{Evaluation Details}
\label{app:evaluation}

\subsection{Baseline provenance and inference protocol}

The corpus-controlled comparison uses the Relation252K-trained CrossPair
checkpoint.  RelationAdapter~\citep{relationadapter} uses its released
Relation252K checkpoint.  Following the provenance statement in the main paper,
the released LoRWeB~\citep{lorweb} comparison is categorized as using
Relation252K together with additional training data.
The Qwen-Image-Edit-2511~\citep{qwenimage} row uses the frozen base editor
without CrossPair LoRA weights.  The \dataset checkpoint is added only to the
OOD panel.

The methods retain their public native generation interfaces, including
method-specific canvas construction and output extraction.  ``Same output
resolution'' in the main comparison refers to the common evaluation image size:
all metric inputs use the same 512-pixel maximum long edge after output
extraction.

\subsection{Deterministic image metrics}

\paragraph{CLIP image similarity.}
We use CLIP~\citep{clip} with the Hugging Face checkpoint
{\urlstyle{tt}\nolinkurl{CLIP-ViT-H-14-laion2B-s32B-b79K}}.  The checkpoint's processor is
applied to each RGB image and image features are \(L_2\)-normalized.
\(\mathrm{CLIP\text{-}I}_{GT}\) is the cosine similarity between generated and
target features, while \(\mathrm{CLIP\text{-}I}_{src}\) compares the generation
with the query source.

\paragraph{CLIP directional similarity.}
For the first demonstration pair (the only pair when \(K=1\)), we compute
\[
\begin{aligned}
  d_r &= \operatorname{norm}\!\left(f(\rt)-f(\rs)\right),\\
  d_q &= \operatorname{norm}\!\left(f(\hat\qt)-f(\qs)\right),
\end{aligned}
\]
and report \(d_r^\top d_q\).  The implementation also supports averaging
multiple demonstration deltas, but this option is not used for the \(K=1\)
experiments.

\paragraph{LPIPS.}
LPIPS~\citep{lpips} uses the AlexNet backbone.  RGB images are mapped from \([0,255]\) to
\([-1,1]\).  Generated images are resized to the target dimensions with Lanczos
interpolation before evaluation.  The evaluation command defaults to a
512-pixel maximum long edge for both generated and target images; the reported
launchers explicitly set this value.

\paragraph{Aggregation.}
The evaluator writes per-sample scores, per-task means, and an overall micro
mean in which every valid query contributes one value.  The main table uses
this query-level micro average, not a transformation-level macro average.  No
confidence interval or significance claim is made for deterministic metrics.

\subsection{Blinded GPT-5.5 ranking}

We use {\urlstyle{tt}\nolinkurl{gpt-5.5-xhigh}} through an OpenAI-compatible Responses API.
For each query, the request contains the reference before, reference after,
query source, and all candidate outputs.  The target image and method names are
never supplied.  Candidate order is shuffled once per query using seed
\(20260724+i\times1009\), where \(i\) is the metadata index.  There is one
request per query ({\urlstyle{tt}\nolinkurl{repeats=1}}), with four candidates in-domain and five
candidates OOD.

Before transmission, every image is EXIF-transposed, converted to RGB,
downscaled with Lanczos to a maximum side of 768 pixels, and JPEG-encoded at
quality 90.  Temperature is omitted, the timeout is 240 seconds, and failed
calls are retried up to three times.  In-domain uses two workers and a
1,600-token response limit; OOD uses four workers and a 2,000-token limit.
Workers affect throughput only.  Returned JSON is accepted only if every
ranking contains every anonymized candidate exactly once without ties.

For each criterion we report Rank~1--\(K\) percentages and mean rank.
Percentages sum to \(100\%\) within a method.  Rank-1 confidence intervals use
1,000 query-level bootstrap resamples and the 2.5/97.5 percentiles.
We additionally export Rank-1 frequency by display position to audit residual
position bias.  The evaluation artifact retains the random candidate mapping,
raw request/response JSON, parsed rankings, failure log, and resume manifest.
The completed common intersections contain 1,091 in-domain queries and 428 OOD
queries, each judged once.  The archived summary files were finalized on July
26 and July 27, 2026, respectively.

\section{DINO Relation Diagnostics and Ablations}
\label{app:diagnostics}

Training logs store the unweighted relation loss, prediction--\allowbreak demonstration
cosine, target--\allowbreak demonstration cosine, prediction--\allowbreak target cosine, change-weight
entropy and maximum weight, pooled-vector norms, and the weighted
auxiliary-to-primary output-gradient norm ratio and cosine.  Debug decoding can
save source/prediction/target panels every 200 steps.  These values diagnose
whether the auxiliary objective is active; they are not treated as image-quality
metrics.

\end{document}